\documentclass[10pt,twocolumn,letterpaper]{article}

\usepackage{cvpr}
\usepackage{times}
\usepackage{epsfig}
\usepackage{graphicx}
\usepackage{amsmath}
\usepackage{amssymb}
\usepackage{gensymb}
\usepackage{makecell}

\usepackage[pagebackref=true,breaklinks=true,letterpaper=true,colorlinks,bookmarks=false]{hyperref}

\cvprfinalcopy 

\def\cvprPaperID{3233} 

\begin{document}

\title{Towards Social Artificial Intelligence: \\Nonverbal Social Signal Prediction in A Triadic Interaction\thanks{Website: \url{http://domedb.perception.cs.cmu.edu/ssp}}}

\author{Hanbyul Joo$^{1}$ \quad   Tomas Simon$^{1}$ \quad Mina Cikara$^2$  \quad Yaser Sheikh$^{1}$\\
$^1$Carnegie Mellon University \, $^2$Harvard University \\
{\tt\small \{hanbyulj, tsimon, yaser\}@cs.cmu.edu, mcikara@fas.harvard.edu}
}

\maketitle

\begin{abstract}
We present a new research task and a dataset to understand human social interactions via computational methods, to ultimately endow machines with the ability to encode and decode a broad channel of social signals humans use. This research direction is essential to make a machine that genuinely communicates with humans, which we call Social Artificial Intelligence. We first formulate the ``social signal prediction'' problem as a way to model the dynamics of social signals exchanged among interacting individuals in a data-driven way. We then present a new 3D motion capture dataset to explore this problem, where the broad spectrum of social signals (3D body, face, and hand motions) are captured in a triadic social interaction scenario. Baseline approaches to predict speaking status, social formation, and body gestures of interacting individuals are presented in the defined social prediction framework. 
\end{abstract}

\section{Introduction}
Consider how humans communicate---we use language, voice, facial expressions, and body gestures to convey our thoughts, emotions, and intentions. Such social signals that encode the messages of an individual are then sensed, decoded, and finally interpreted by communication partners. Notably, the use of all these channels is important in social interactions, where subtle meanings are transmitted via the combination of such signals. Endowing machines with the similar social interaction ability that encodes and decodes a broad spectrum of these social signals is an essential goal of Artificial Intelligence (AI) to make them to effectively cooperate with humans. We use the term, \emph{Social Artificial Intelligence (AI)}, to refer to the machine with such ability. A way to build the \emph{Social AI} would be to encode all the rules that humans observe during social communication~\cite{cassell1994animated}. Unfortunately, nonverbal interaction is still poorly understood despite its importance in social communication~\cite{Mehrabian67,Mehrabian81,Birdwhistell-1970}, making it hard to formalize rules about how to understand and use social signals. Interestingly, we have recently witnessed a great achievement in Natural Language Processing showing the potential to make machines ``freely" communicate with humans using language~\cite{young2018recent}. This success has been led by data-driven approaches leveraging large-scale language datasets and a powerful modeling tool, deep neural networks, to automatically learn the patterns of human verbal communication. Remarkably, these achievements have not made extensive use of the prior knowledge about grammar and the structure of languages that linguists have accumulated over centuries. Motivated by this, we hypothesize that a similar approach can be applicable in modeling nonverbal communication. 

However, there exists a fundamental difficulty in building a data-driven nonverbal communication model: the data is extremely rare. In the verbal language domain, words contain the full expressive power to record verbal signals by the composition of a handful of discrete symbols, and there already exist millions of articles, dialogues, and speech audio on the Internet which are readily usable for data-driven methods. However, for non-verbal signals, how to ``record" or collect these signals is uncertain. Imagine a situation where a group of people are communicating in our daily life. The positions and orientations of individuals, their body gestures, gaze, and facial expressions (among others) are the data we are interested in. Notably, these social signals emitted from all people in the group (i.e., the signals from senders and receivers) need to be collected simultaneously to study their correlation and causality. Although there also exist millions of videos where our daily activities---including social interactions---are captured on the Internet, these raw videos cannot be directly used to understand the rules of non-verbal interactions because we have to extract all semantic visual cues (relative location, face, body pose, and so on) from the raw pixels. 

\begin{figure}[t]
	\centering
	\includegraphics[width=\linewidth]{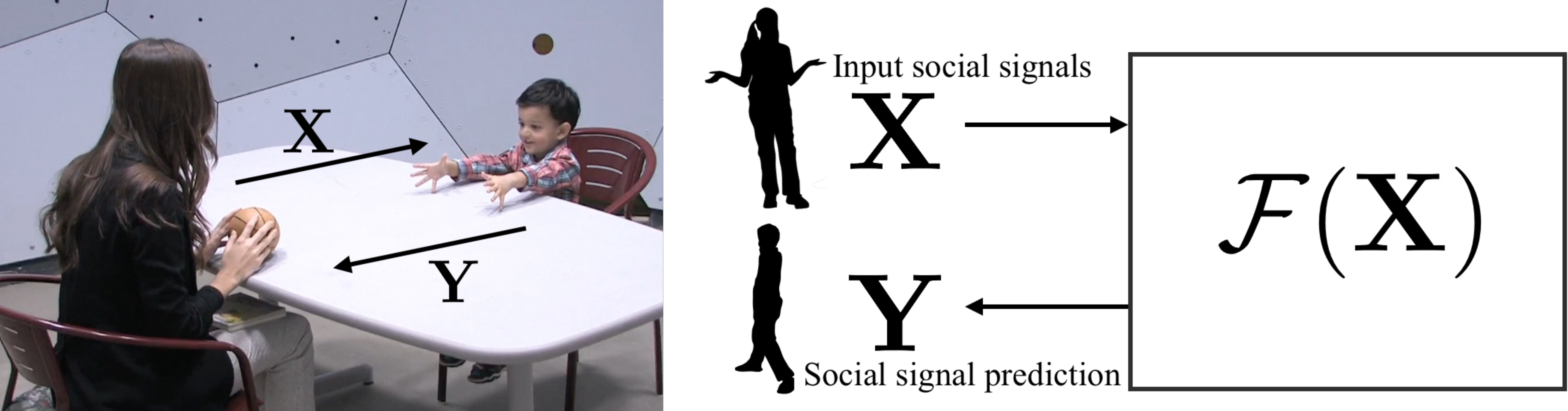}
	\caption{We aim to learn the dynamics between the input signals $\mathbf{X}$ that an individual receives and the output social signals $\mathbf{Y}$ the individual emits. The goal of social signal prediction is to regress a function $\mathbf{Y}{=}\mathcal{F}(\mathbf{X})$ in a data-driven manner.}
	\label{fig:ssp_intro}
\end{figure}

In this paper, we introduce a new research domain with the long-term aim to build machines that can genuinely interact with humans using non-verbal signals. The key idea of this problem is to define the social interaction as a signal prediction problem among individuals, rather than finding the mapping between the social signals and their semantics (e.g., as in affective computing or emotion recognition~\cite{picard1997affective, picard2003affective,poria2017review, ekman1969}). We formulate that humans are communicating by receiving a set of social signals from others and emitting response signals as output, which are again directed to others as inputs (illustrated in Fig.~\ref{fig:ssp_intro}). Thus, we hypothesize that the social behavior ability can be learned by modeling the dynamics between these signal flows. The advantages of this approach are: (1) we can tackle the problem by investigating the objectively measurable social signal data avoiding annotating the underlying meanings of these signals and (2) it enables us to model subtle details of social communication by considering the original continuous and high-dimensional signal space.

Importantly, we aim to include as many channels of signals as possible to study the correlation between various channels of social signals, including facial expressions, body gestures, body proximity, and body orientations. Since collecting these signals in-the-wild is hard due to the limits of signal measurement technology (e.g., motion capture) as well as the large variety of interaction contexts, we collect a large-scale dataset in a studio setup~\cite{joo2015panoptic, joo2017panoptic} where these signals are markerlessly measured for naturally interacting people. Our dataset includes about 180 sequences in a specific triadic social interaction scenario, where 120 subjects participated. Directly investigating the dynamics of the full spectrum of social signals is challenging due to the high complexity of the motion space, requiring a much larger scale of data. We thus simplify the problem by focusing on predicting lower dimensional, yet important, output signals---speaking status and social formations---emitted by the target person, while still considering broader channels including body motion and facial expressions as input. We found that this approach still provides an important opportunity to computationally study various channels of interpersonal social signals. Results on our baseline social signal prediction models, implemented by neural networks, demonstrate the strong predictive property among social signals and also allow us to quantitatively compare the relative influence among them. We further discuss more challenging prediction tasks, including body gesture prediction and entire visual social signal prediction.

\section{Background}

\textbf{Behavioral Science:}
Due to its importance in social communication, nonverbal cues have received significant attention in psychology and behavioral science. Work in this area is often categorized into diverse subfields including Proxemics, Kinesics, Vocalics, and Haptics~\cite{Moore13}. In our work, we focus only on Proxemics and Kinesics, which are closely related to visual cues. Hall first introduced the concept of Proxemics to describe how humans use their personal space during communications~\cite{Hall66}, and Kendon studied the spatial formations and orientations established when multiple people communicate in a common space (named F-formation)~\cite{kendon90}. Facial expression in particular has received lots of attention by researchers since the pioneering work of Charles Darwin~\cite{Darwin-1872}. Ekman studied the relation between emotions and facial expressions, and presented the Facial Action Coding System (FACS), a system to describe facial expressions using combinations of atomic units (Action Units)~\cite{ekman1977facial}. Since then, this system remains the de-facto standard to annotate and measure facial expressions, and has had a broad impact in across many fields. Compared to the face, body gestures remain relatively unexplored, even though research has substantiated the importance of body language in communication~\cite{Gelder09, Moore13, Meeren-2005, Aviezer-2012}. Despite the efforts of many researchers in diverse fields, little progress has been made in understanding nonverbal communications, and the approaches proposed several decades ago are still the most widely used methods available~\cite{Moore13}. In particular in this field, researchers have been using manual behavioral coding procedures that are several decades old; however, most people are moving away from manual behavioral coding towards automated coding.

\textbf{Social Signal Processing:}
There has been increasing interest in studying nonverbal communication using computational methods~\cite{vinciarelli2009social}. Analyzing facial expression has been a core focus in the vision community ~\cite{ChuDC13, Torre15, shan2009facial}. Many other methods to automatically detect social cues from photos and videos have also been proposed, including F-formation detection~\cite{setti2015f}, recognizing proxemics from photos~\cite{yang2012recognizing}, detecting attention~\cite{Fathi-2012}, recognizing emotions by body pose\cite{schindler2008recognizing}, and detecting social saliency~\cite{park20123d}. Affective computing fields have been growing up rapidly, where computer vision and other sensor measurements are used with machine learning technique to understand human's emotion, social behavior, and roles~\cite{picard1997affective}. 

\textbf{Forecasting human motion:}
Predicting or forecasting human motion is an emerging area in computer vision and machine learning. Researchers propose approaches for predicting pedestrian's future trajectory~\cite{kitani2012activity} or forecasting human interaction in dyadic situations~\cite{huang2014action}. More recently, deep neural networks are used to predict future 3D poses from motion capture data~\cite{mnih2012conditional, Fragkiadaki_2015_ICCV, jain2016structural}, but they focus on periodic motions such as walk cycles. Recent work tries to forecast human body motion in the 2D image domain~\cite{walker2016uncertain, villegas2017learning}. A few approaches address trajectory prediction in social situations~\cite{helbing1995social, alahi2016social, gupta2018social}. 


\textbf{Social Signal Dataset:}
How to measure and collect nonverbal signal data is important to pursue a data-driven approach for our goal. However, only few datasets contain socially interacting group motion~\cite{SALSA-15, mccowan2005ami, lepri2012connecting, rehg2013decoding}. The scenes in these datasets are often in a table setup, limiting free body movement and capturing the upper-body only. There are datasets that capture more freely moving multiple people (e.g., cocktail party)~\cite{Zen-10, Cristani-11, farenzena2009social}, but these only contain location and orientation measurements for the people, introduced to study the social formation only. Datasets providing rich 3D body motion information captured with motion capture techniques exist, but they contain single subjects' motion only~\cite{gross2001cmu, h36m_pami, sigal2010humaneva}. More recently, however, full body motion data of interacting groups using a large number of camera system was proposed for social interaction capture~\cite{joo2015panoptic,joo2017panoptic}. This work shows a potential in collecting a large scale social interaction data without the usual issues caused by wearing a motion capture suit and markers.

\textbf{Measuring Kinesic Signals:} Detecting human bodies and keypoints in images has advanced greatly in computer vision. There exist publicly available 2D face keypoint detectors~\cite{baltruvsaitis2016openface}, body pose detectors~\cite{cao2017realtime, Wei2016, Newell-16}, and hand pose detectors~\cite{simon2017hand}. 3D motion can be obtained by markerless motion capture in a multiview setup~\cite{Gall-09,Liu-2013,Elhayek-15, joo2017panoptic, joo2018}, by RGB-D cameras~\cite{Shotton2011,Baak2011}, or even by monocular cameras~\cite{Ramakrishna2012,Bogo2016,martinez2017simple,zhou2017towards,Moreno-noguer2017,mehta2017monocular}. Recently, methods to capture both body and hands have also been introduced~\cite{MANO:SIGGRAPHASIA:2017,joo2018}.

\section{Social Signal Prediction}
The objective of \emph{Social Signal Prediction} is to predict the behavioral cues of a target person in a social situation by using the cues from communication partners as input (see Fig.~\ref{fig:ssp_intro}). We hypothesize that the target person's behavior is correlated with the behavioral cues of other individuals. For example, the location and orientation of the target person should be strongly affected by the position of conversational partners (known as Proxemics~\cite{Hall66} and F-formation~\cite{kendon90}), and the gaze direction, body gestures, and facial expressions of the target person should also be ``conditioned" by the behaviors of the conversational partners. In the social signal prediction task, we model this conditional distribution among interacting subjects, to ultimately teach a robot how to behave in a similar social situation driven by the behavior of communication partners. There exist cases where the correlation of the social signals among subjects is strong, such as hand-shaking or greeting (waving hands or bowing). But in most of the cases, the correlation is implicit---there exist no specific rules on how to behave given other people's behavior, which makes it hard to manually define the rules. In our approach, we tackle this problem in a data-driven manner, by automatically learning the conditional distributions using a large scale multimodal social signals corpus.  

We first conceptually formulate the social signal prediction problem here, and a specific implementation focusing on the Haggling scenario is described in the next section. Let us denote ``all types of signals'' that the target person received in a social situation at time $t$ as $\mathbf{X}(t)$. Thus $\mathbf{X}(t)$ includes the social signals from other individuals---body gestures, facial expression, body position, voice tones, verbal languages---and also other contextual factors such as the space where the conversation is performed or other visible objects which may affect the behavior of the target person (e.g., sounds or objects in the environment may attract the attention of the person). We divide the input signal $\mathbf{X}(t)$ into two parts, the signals from the conversational partners, $\mathbf{X}_c(t)$, and signals from other sources (e.g., objects, environment, and other human subjects not interacting with the target person), $\mathbf{X}_e(t)$.  Thus,
\begin{equation}
\mathbf{X} (t) = \{  \mathbf{X}_c (t), \mathbf{X}_e (t)\}.
\end{equation}
The term $\mathbf{X}_c (t)$ may contain the social signals from multiple people and we denote the signals from each subject separately:
\begin{equation}
\mathbf{X}_c (t) = \{ \mathbf{X}_c^i (t) \}_{i=1}^N,
\end{equation}
where $\mathbf{X}^i_c (t)$ are the signals from the $i$-th conversational partner in the social interaction and $N$ is the total number of partners. 
We also denote the signals emitted by the target person at time $t$ in the social situation as $\mathbf{Y} (t)$.
Then, the goal of social signal prediction is to find a function $\mathcal{F}$ which takes $\mathbf{X}$ as input and produces $\mathbf{Y}$ as output to mimic the behavior of the target person in the social situation:
\begin{equation}
\mathbf{Y} (t+1) = \mathcal{F} \left( \mathbf{X} (t_0:t), \mathbf{Y} (t_0:t) \right),
\label{equation:socialPrediction_1}
\end{equation}
where $t_0:t$ represents a range of time from $t_0$ to $t$ affecting the current behavior of the target person. Note that we define the function $\mathcal{F}$ to take the history of the target person's own signals $\mathbf{Y} (t_0:t)$, and the function predicts the immediate future motion (or response) of the target individual. Intuitively, this formulation models the human behavior as a function that represents the dynamics among social signals the target person is receiving and emitting. The function can be defined for a specific individual, representing the personal behavior of the target person encoding characteristics, like physical attributes or culture of the target. Based on that, different individuals may behave differently. If the function is regressed by the data from many people, then we hypothesize that the function produces more general and common social behaviors, where the individual specific behaviors are averaged out.

Previous approaches can be considered as subsets of this model. For example, conversational agents (or chatbots) using natural language only can be represented as:
\begin{equation}
\mathbf{Y}_v (t+1) =  \mathcal{F} \left( \mathbf{X}_v (t_0:t) \right),
\end{equation}
where $\mathbf{Y}_v$ and $\mathbf{X}_v$ represents only verbal signals. The human motion forecasting studied in computer vision and graphics~\cite{Fragkiadaki_2015_ICCV, jain2016structural} can be considered as:
\begin{equation}
\mathbf{Y}_n (t+1) =  \mathcal{F} \left( \mathbf{Y}_n (t_0:t) \right),
\end{equation}
where $\mathbf{Y}_n$ represents nonverbal body motion. Note that in this task, there is no social interaction modeling, and the prediction is only for an individual using the individual's own previous signals. 

In our work, we focus on nonverbal social signal prediction in a triadic social interaction scenario:
\begin{equation}
\mathbf{Y} ( t_0:t ) =  \mathcal{F} \left( \mathbf{X}_c^1 (t_0:t), \mathbf{X}_c^2 (t_0:t) \right),
\label{equation:F_ours}
\end{equation}
where we predict the social signals of the target given signals of the two other people during the same window of time. In particular, we consider diverse input and output social signals to investigate their dynamics and correlations. 

\section{Triadic Interaction Dataset with Full-spectrum Social Signal Measurements}
\label{chapter:dataset}
Availability of a large-scale dataset is essential to computationally investigate the nonverbal communication in a data-driven manner. Despite existing datasets that provide measurements for human motion and behaviors~\cite{carletta2005ami, Lepri-12, Zen-10,Cristani-11, SALSA-15, h36m_pami}, there is no dataset that satisfies the following core requirements for understanding nonverbal human behaviors: (1) capturing 3D full body motion with a broad spectrum of nonverbal cues (including face, body, and hands); (2) capturing signals of naturally interacting groups (more than two people to include attention switching); and (3) collecting the data at scale. The limited availability of datasets motivates us to build a new dataset that contains social interactions among hundreds of interacting groups with a broad spectrum of 3D body motion measurements. The key properties of our dataset are as follows:
\begin{itemize}
	\setlength\itemsep{0em}
	\item Naturally interacting multiple people in a negotiation game scenario, where the game is carefully designed to induce natural and spontaneous interaction
	\item No behavioral restriction is instructed to participants during the capture
	\item A broad spectrum of social signals, including the motion from faces, bodies, and hands, are measured using a state-of-the art markerless motion capture system~\cite{joo2015panoptic, joo2017panoptic}
	\item Multiple synchronized modalities, including RGB videos from over 500 views, depth maps from 10 RGB+D sensors, and sound from 23 microphones  
	\item Voice signals of individuals are recorded via wireless and wired microphones, and speaking status and timing of each subject are manually annotated
	\item 3D point clouds are provided by fusing the depth maps from 10 RGB+D sensors
\end{itemize}

Our dataset provides a new opportunity to investigate the dynamics of various interpersonal nonverbal behavioral cues emerging in social situations. Our dataset is captured under a university-approved IRB protocol\footnote{IRBSTUDY2015\_00000478} and publicly released for research purposes.


\begin{figure}
	\centering
	\includegraphics[trim=300 0 0 0,clip,width=0.49\linewidth]{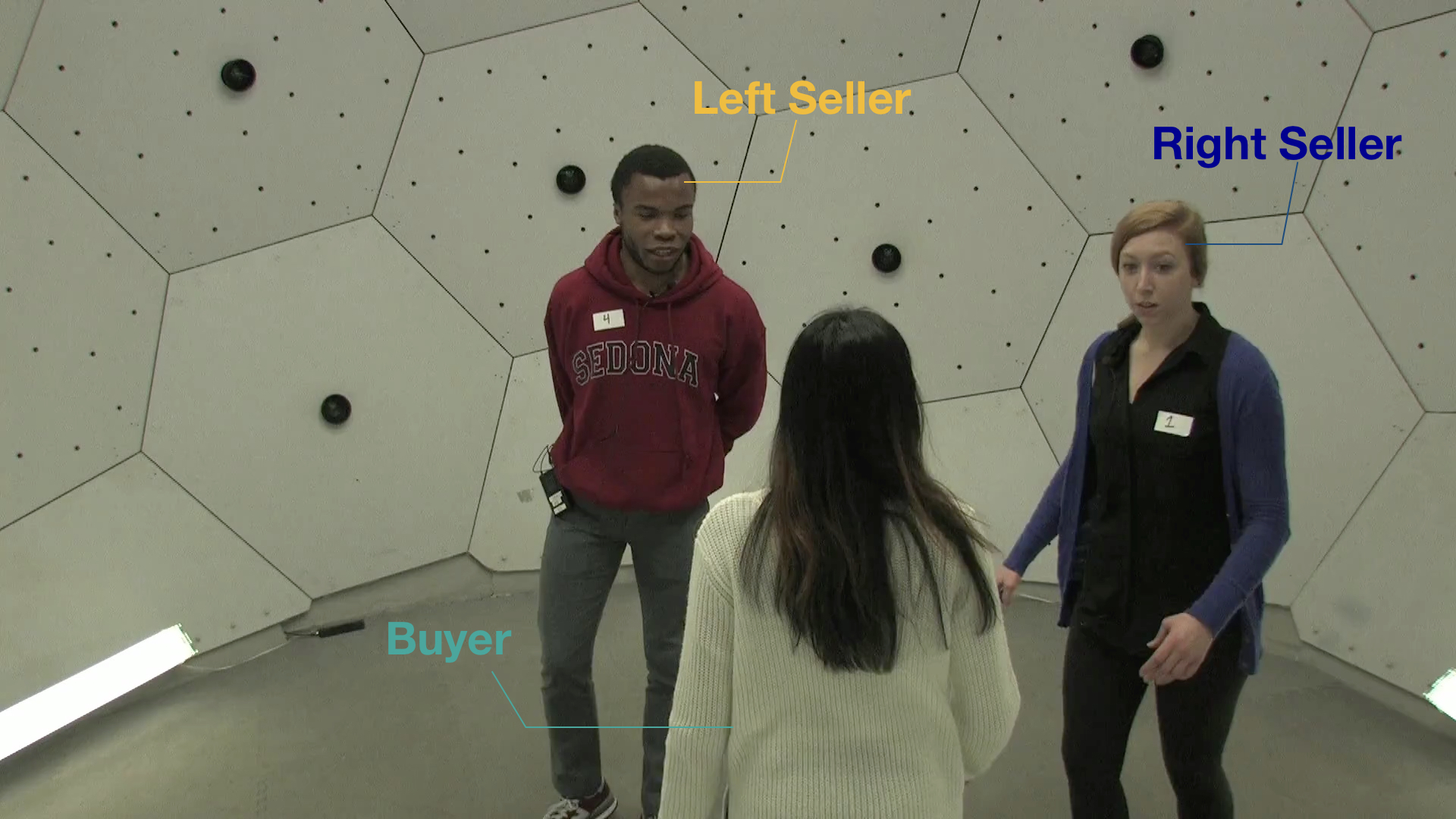}
	\includegraphics[trim=250 230 250 0,clip,height=0.30\linewidth]{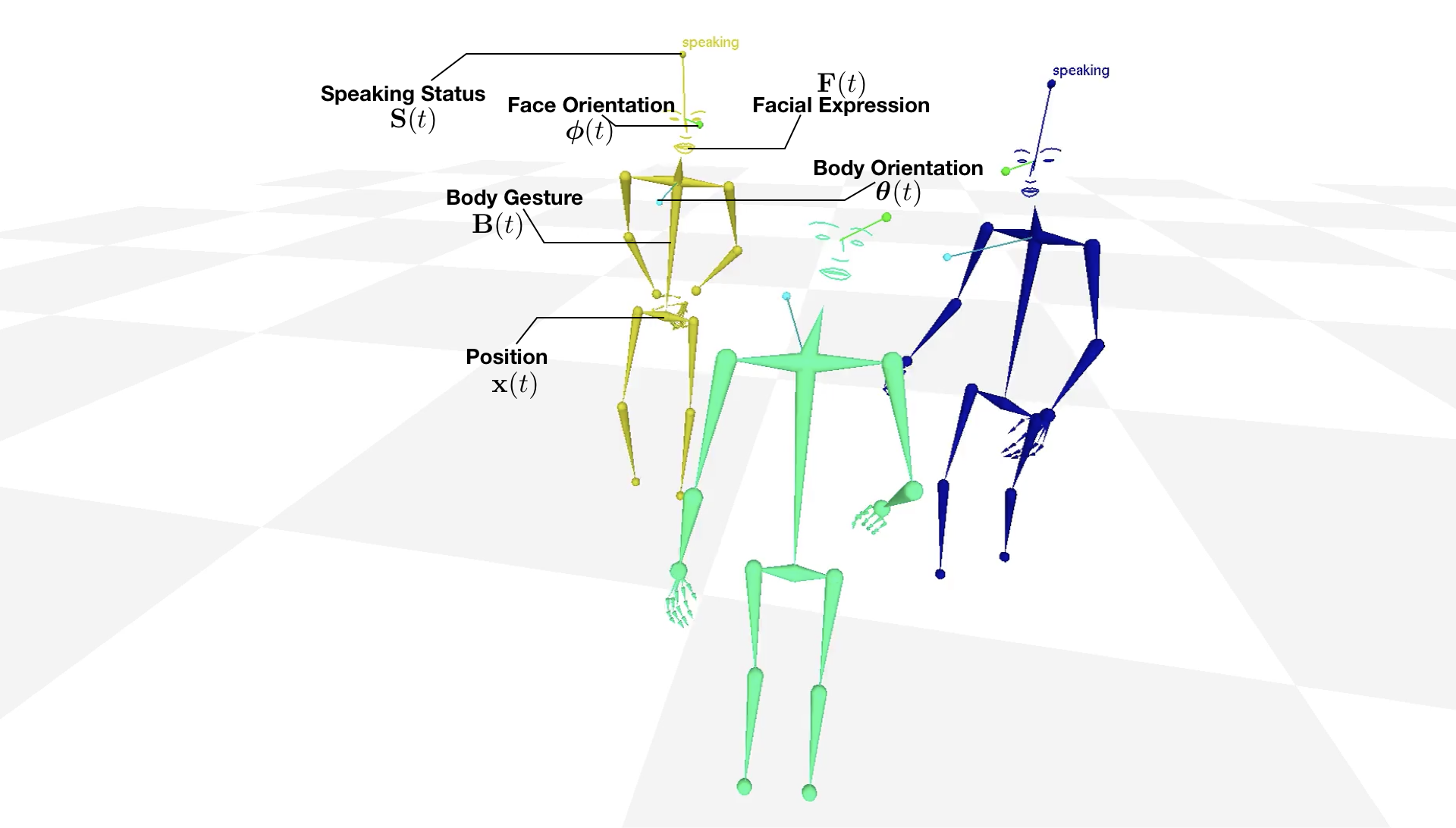}
	\caption{An example of the haggling sequence. (Left) an example scene showing two sellers and one buyer. (Right) Reconstructed 3D social signals showing the 3D body, 3D face, and 3D hand motion. 3D normal direction from faces and bodies are also computed, and speaking status of each individual, manually annotated, is also visualized here.} 
	\label{fig:hagglign_ex}
\end{figure}


\subsection{The Haggling Game Protocol}

To evoke natural interactions, we involved participants in a social game named the \emph{Haggling} game. We invent this game to simulate a haggling situation among two sellers and a buyer. The triadic interaction is chosen to include interesting social behaviors such as turn taking and attention changes, which are missing in previous dyadic interaction datasets~\cite{rehg2013decoding}. During the game, two sellers are promoting their own products for selling, and a buyer makes a decision about which product he/she buys between the two. The game lasts for a minute, and the seller who has sold his/her product is awarded $\$5$. To maximize the influence of each seller's behavior on the buyer's decision-making, the items assigned to sellers are similar products with slightly different properties. Example scenes are shown in Figure~\ref{fig:hagglign_ex} and our supplementary video. See the supplementary material for the detailed game protocol.

\subsection{Measuring Social Signals and Notation}
\label{section:Notation}
We use the Panoptic Studio to reconstruct 3D anatomical keypoints of multiple interacting people~\cite{joo2015panoptic, joo2017panoptic}. As a key advantage, the method does not require attaching sensors or markers on subject's body, and no behavior restrictions nor initialization poses are needed from the subjects. As output, the system produces 3D body motion $\mathbf{B}(t)$ and 3D face motion $\mathbf{F}(t)$ for each individual at each time $t$\footnote{The system also produces 3D hand motion, but we do not use this measurement in this paper due to the occasional failures in challenging hand motions (e.g., when both hands are close to each other). However, believing that these are still important cues, we release the reconstruction results for future work.}. We also denote the global position of the body as $\mathbf{X}(t)$. From these measurements, we additionally compute the body orientation $\boldsymbol{\theta}(t)$ and face orientation $\boldsymbol{\phi}(t)$ by finding the 3D normal direction of torso and face, respectively. We describe the detailed representation of the measurements below. See Fig.~\ref{fig:hagglign_ex} for the visualization of measured signals.

\paragraph{Body Motion:} We follow the body motion representation of the work of Holden et al.~\cite{holden2016deep}, representing a body gesture at a frame as a 73-dimensional vector, $\mathbf{B}(t) \in \mathbb{R}^{73}$. This representation is based on the skeletal structure of CMU Mocap dataset~\cite{gross2001cmu} with 21 joints (63 dimensions), along with the projection of the root joint (the center of the hip joints) on the floor plane (3 dimensions), the relative body locations and orientations represented by the velocity values of the root (3 dimensions), and footstep signals (4 dimensions). The orientations are computed only on the $x$-$z$ plane with respect to the $y$-axis, and the location and orientation represent the changes from the previous frame rather than the absolute values, following the previous work~\cite{jain2016structural, holden2016deep}. In particular, the first 63 dimensions of $\mathbf{B}(t)$ represents the body motion in person-centric coordinates, where the root joint is at the origin and torso is facing the $z$ direction. We perform a retargeting process to convert our original 3D motion data from the Panoptic Studio, where the skeleton definition is the same as COCO dataset~\cite{coco-14}, to this body motion representation with a fixed body scale. Thus, in our final motion representation, individual specific cues such as heights or lengths of limbs are removed and only motion cues are kept.
\paragraph{Face Motion:} For the face motion signal, we first fit the deformable face model of \cite{cao2014facewarehouse} and use the initial 5 dimensions of the facial expression parameters, because we found the remaining dimensions have an almost negligible impact on our reconstruction quality. Note that the face expression parameters in \cite{cao2014facewarehouse} are sorted by their influence by construction and the initial components have more impact in expressing facial motion. To this end, face motion at a time instance is represented by a 5-dimensional vector, $\mathbf{F}(t) \in \mathbb{R}^{5}$ (See the supplementary material for the visualization of these facial expressions). Here, we also do not include individual-specific information (the face shape parameters varying for individuals) and only motion cues are kept.

\paragraph{Position and Orientation:} For the global position $\mathbf{x}(t)$ of each individual, we use the coordinate of the root joint of the body, ignoring the values in $y$ axis, and thus $\mathbf{x}(t) \in \mathbb{R}^2$.
We use a 2D unit vector to represent body orientations $\boldsymbol\theta (t) \in \mathbb {R}^2$ and face orientation $\boldsymbol\phi (t) \in \mathbb{R}^2$, defined on the $x$-$z$ plane ignoring the values in $y$ axis. Note that we use unit vectors rather than angle representation, because the angle representation has a discontinuity issue when wrapping around $2\phi$ and $-2\phi$. In contrast to the relative location and orientation represented in the part of body motion $\mathbf{B}(t)$, these $\mathbf{x}(t)$, $\boldsymbol\theta (t)$, and $\boldsymbol\phi (t)$ represent the values in the global coordinate, which are used to model social formation. In summary, the status of an individual at a frame in social formation prediction is represented by a 6-dimensional vector, $[\mathbf{x}(t)^{\top}, \boldsymbol{\theta}(t)^{\top}, \boldsymbol{\phi}(t)^{\top} ]^{\top} \in \mathbb{R}^6$.

\paragraph{Speaking Status:} The voice data $\mathbf{V}(t)$ of each individual is also recorded by wireless microphones assigned to each individual. From the audio signal, we manually annotate a binary speaking label $\mathbf{S}(t) \in \{0,1\}$ describing whether the target subject is speaking (labelled as $1$) or not speaking (labelled as $0$) at time $t$.\\
\mbox{ }\\
By leveraging these various behavioral cues measured in the Haggling scenes, we model the dynamics of these signals in a triadic interaction. The objective of our direction is to regress the function defined in Equation~\ref{equation:F_ours}. To further constrain the problem we assume that the target person is the seller positioned on the left side of the buyer\footnote{To simplify the problem, particularly for the formation prediction.}, and as input we use the social signals of the buyer ($\mathbf{X}^1$) and the other seller ($\mathbf{X}^2$). Based on our social signal measurements, the input and output of the function are represented as,
\begin{equation}
\begin{gathered}
\mathbf{Y} = [ \mathbf{x}^0, \boldsymbol{\theta}^0, \boldsymbol{\phi}^0, \mathbf{B}^0, \mathbf{F}^0, \mathbf{S}^0 ],\\
\mathbf{X}^1 = [ \mathbf{x}^1, \boldsymbol{\theta}^1, \boldsymbol{\phi}^1, \mathbf{B}^1, \mathbf{F}^1, \mathbf{S}^1 ],\\
\mathbf{X}^2 = [ \mathbf{x}^2, \boldsymbol{\theta}^2, \boldsymbol{\phi}^2, \mathbf{B}^2, \mathbf{F}^2, \mathbf{S}^2 ],
\end{gathered}
\end{equation}
where we use the superscript 0 to denote the social signals of the target subject (the output of social signal prediction).

\section{Social Signal Prediction in Haggling Scenes}
We use our Haggling scenario as an example problem of social signal prediction to computationally model triadic interaction. In this section, we specifically define the input and output signals used in our modeling, and then present three social signal predicting problems, predicting speaking status, predicting social formation, and predicting body gestures (Kinesic signals). Note that we focus on estimating the target person's concurrent signals by taking other individuals' signals as input as defined in Equation~\ref{equation:F_ours} to simplify the problem, rather than forecasting the future signals. See the supplementary material for the implementation details.

\subsection{Predicting Speaking}
\label{subsection:ssp_pred_speak}
We predict whether the target subject is currently speaking or not, denoted by $\mathbf{S}^0$.  This is a binary classification task and can be trained with a Cross Entropy loss. We first study the correlation between the speaking signal of the target person, $\mathbf{S}^0$, and the person's own social signals, either body motion $\mathbf{B}^0$ or facial motion $\mathbf{F}^0$, or both. We expect this correlation is stronger than the link across individuals. 
Formally, a function $\mathcal{F}_{B0\rightarrow S0}$ takes the target person's own body motion $\mathbf{B}^0(t_0:t)$ to predict the speaking signal:
\begin{gather}	
\mathbf{S}^0(t_0:t) = \mathcal{F}_{B0\rightarrow S0} \left( \mathbf{B}^0(t_0:t) \right),
\label{eq:speaking_0}
\end{gather}
and similarly,
\begin{gather}	
\mathbf{S}^0(t_0:t) = \mathcal{F}_{F0\rightarrow S0} \left( \mathbf{F}^0(t_0:t) \right),\\
\mathbf{S}^0(t_0:t) = \mathcal{F}_{(F0, B0)\rightarrow S0} \left( \mathbf{F}^0(t_0:t) , \mathbf{B}^0(t_0:t)\right),
\label{eq:speaking_0_facebody}
\end{gather}
where $\mathcal{F}_{F0\rightarrow S0}$ takes the target person's own face motion, and $\mathcal{F}_{(F0, B0)\rightarrow S0}$ takes both face and body cues. We compare the performance of these functions with the function that takes the signals from a communication partner, the other seller:
\begin{gather}	
\mathbf{S}^0(t_0:t) = \mathcal{F}_{B2\rightarrow S0} \left( \mathbf{B}^2(t_0:t) \right),\\
\mathbf{S}^0(t_0:t) = \mathcal{F}_{F2\rightarrow S0} \left( \mathbf{F}^2(t_0:t) \right),\\
\mathbf{S}^0(t_0:t) = \mathcal{F}_{(F2, B2)\rightarrow S0} \left( \mathbf{F}^2(t_0:t), \mathbf{B}^2(t_0:t) \right),
\label{eq:speaking_1}
\end{gather}
where the functions use body cues, face cues, and both cues, respectively. 

This framework enables us to quantitatively investigate the link among social signals across individuals. For example, we may easily hypothesize that there exists a strong correlation between the signals from the same individual (e.g., speaking and facial motion of the target person), while the correlation between the signals across different individuals (e.g., speaking of the target person and body motion of another person) may be considered as weak. By comparing their performances, we verify there still exists strong links among these signals exchanged across subjects.

\subsection{Predicting Social Formations}
We predict the location and orientations of the target person, denoted by $\mathbf{Y}_p = [{\mathbf{x}^0}^{\top}, {\boldsymbol{\theta}^{0}}^{\top}, {\boldsymbol{\phi}^{0}}^{\top} ]^{\top}$, given the same channels of cues from the communication partners. This problem is strongly related to Proxemics~\cite{Hall66} and F-formation~\cite{kendon90}, illustrating how humans use their space in social communications. Formally, 
\begin{gather}	
\mathbf{Y}_p (t_0:t) = \mathcal{F}_p \left( \mathbf{X}_p^1(t_0:t), \mathbf{X}_p^2(t_0:t) \right),
\label{eq:pred_formation}
\end{gather}
where $\mathbf{Y}_p$, $\mathbf{X}_p^1$, and $\mathbf{X}_p^2$ contain global location and orientation signals $[{\mathbf{x}^i}^{\top}, {\boldsymbol{\theta}^{i}}^{\top}, {\boldsymbol{\phi}^{i}}^{\top} ]^{\top}$ (where $i=0,1$, or $2$) for the target subject and others.
Note that we only consider the positions and orientations on the ground plane (in 2D), ignoring the height of the subjects, and thus $\mathbf{Y}_p(t), \mathbf{X}_p^i(t) \in \mathbb{R}^6$. This prediction problem is intended to see whether the machine can learn how to build a social formation to interact with humans~\cite{vazquez2017towards}.

\subsection{Predicting Body Gestures (Kinesic Signals)}
Predicting body motion in social situations (by using other subjects' signals) is challenging, because the correlation among body signals are subtle and less explicit. To study this, we present two baseline approaches here. 

\textbf{By Using Social Formation Only.} The first approach uses only the social formation information of other subjects:
\begin{gather}	
\mathbf{B}^0(t_0:t) = \mathcal{F}_{p\rightarrow B0 } \left( \mathbf{X}_p^1(t_0:t), \mathbf{X}_p^2(t_0:t) \right).
\label{eq:pred_traj2body}
\end{gather}
This is an ill-posed problem with diverse possible solutions, because the formation signals of communication partners barely tells us about the detailed behavior of our target person. Yet, we can consider several required properties of the predicted skeleton. For example, the body location and orientation need to satisfy the social formation property, and when the target person's location is changing the appropriate leg motion needs to be predicted. Intuitively, we expect the predicted skeleton shows a similar social amount of information, location and orientations, as in social formation prediction, but using more complicated structure, body motion. In that sense, we can divide the function $\mathcal{F}_{j0}$ into two stages: predicting a social formation by $\mathcal{F}_p$ described in Equation~\ref{eq:pred_formation} and predicting 3D body motion from the predicted social trajectory $\mathbf{Y}_p (t_0:t)$:
\begin{gather}	
\mathbf{B}^0 (t_0:t) = \mathcal{F}_{p\rightarrow B0 }  \left(   \mathcal{F}_p \left( \mathbf{X}_p^1(t_0:t), \mathbf{X}_p^2(t_0:t) \right) \right) \nonumber \\ 
= \mathcal{F}_{p\rightarrow B0 } \left( \mathbf{Y}_p (t_0:t)  \right),
\label{eq:pred_p2J}
\end{gather}
where $\mathcal{F}_{p\rightarrow B0 }$ is a mapping between the target subject's own social trajectory to body skeleton. Since the trajectory (position and orientations) is a sub-part of the body behavior, we expect the predicted skeleton to contain similar signals as the social trajectory. For the function $\mathcal{F}_{p\rightarrow B0 }$, we follow a similar approach to the work of Holden et al.~\cite{holden2016deep}.

\textbf{By using Body Motions as Input.} We can use the entire body signals of conversational partners as input for our function:
\begin{gather}	
\mathbf{B}^0 (t_0:t) = \mathcal{F}_{(B1,B2)\rightarrow B0} \left( \mathbf{B}^1 (t_0:t), \mathbf{B}^2 (t_0:t) \right) .
\label{eq:pred_body2body}
\end{gather}
In this particular example, we expect ``better" prediction quality than the previous baseline by using other subject's body motions as a cue to determine the target person's body motion. We found that this method shows more diverse upper body motion, responding to the motions of other subjects. To this end, we present a hybrid method combining the upper body prediction results of this method to the root and leg motions of the previous method.

\section{Results}
In this section, we show experimental results for three prediction tasks, predicting speaking status, social formations, and gestures (kinesic signals) from different input sources. The core direction is to explore the existence of correlations of diverse behavioral channels in genuine social communications. 

\subsection{Pre-processing Haggling Data}
Given the measurement data of the Haggling games, we first manually annotate the start and end time of the game, where the start time is decided when the social formation is built and the end time is defined when the social formation is broken. We crop out the motion dataset based on this start and end times, so that we ignore the time while subjects enter and exit the capture space. For each haggling game scene, we also annotate the players' roles in the game, buyer, left-seller, and right-seller, where the left and right are determined in the buyer's viewpoint. In our experiment, we specify that the left seller is our target person and predict the social behavior of these subjects. As described in our method section, we re-target the motion data to a standardized skeleton size to remove size variation from the body skeletons, similar to \cite{holden2016deep}. We also synthesize footstep signals and decouple the body motion from global translation and orientation using the method of \cite{holden2016deep}. For face motion, we fit the Facaewarehouse model~\cite{cao2014facewarehouse} on the 3D keypoints of individual's face, and use the first five facial expression parameters, as described in Section~\ref{section:Notation}. Finally, we divide the dataset into 140 training sets and 40 test sets. However, since there exist sequences where the reconstruction errors are severe for some frames, we select only 79 training sets and 28 testing sets which are manually verified to be error free. We additionally divide all training set into slices with 120 frames with a certain interval (10 frames), and generate about 10K training samples. We also consider a flipped version by considering the ``right person'' as the target in the same group, generating about 20K training samples in total for the training. We standardize all input data so that they have zero mean and unit variance.

\subsection{Speaking Classification}
We predict whether the target person is currently speaking or not by observing other channels of social signals. 

\textbf{Result on Intra-personal Signals.} First, we investigate the performance when the target individual's own social signals are used as input. Three different input sources---facial expressions, body gestures, and both of them---are used to train neural network models respectively. In particular, we use the same neural network architecture for this experiment by keeping the input dimension and network size as the same to make the comparison as fair as possible (See the supplementary material). To train the network with different types of input, we mask out the unused channels in the input with their average, computed in the training set. The prediction accuracies from these input signals are shown in the second column of Table~\ref{table:speaking_class} (labeled as ``Self''). As demonstrated in our result, the social signals from the target individual show strong correlations with the speaking. For example, the facial cue of the target person shows the strongest correlation (about $89\%$ accuracy) with the target person's own speaking status, presumably due to the strong correlation between the lip motion and speaking. The body motion also shows a strong correlation with more than $73\%$ prediction accuracy. The result with both body and face signals, shown in the second row labeled ``Face+Body'' of Table~\ref{table:speaking_class}, is similar to the case that only the face cue is used, and by applying an ablation study we found that this is because the network predominantly uses the face cues over the body cues for the prediction, as shown in the last two rows of the Table~\ref{table:speaking_class_ablation}. More specifically, given the trained model of ``Face+Body'', we mask out the face part or body part in the input data (at ``testing'' time) and evaluate the performances. The accuracy after removing the body part is similar to the original performance, meaning that the trained network is less dependent on body cues, while there exists a much larger drop if the face part is removed. 

\textbf{Result on Inter-personal Signals.}
A more interesting experiment is investigating the performance by using the other seller's social signals as input to predict the target person's speaking status. Similarly, three different input sources are considered, and the results are shown in the second column of the Table~\ref{table:speaking_class}.  The result shows that there exists a strong link between interpersonal social signals. The other seller's facial motion shows strong predictive power for the target person's speaking status, with accuracy higher than when using only the target person's own body signals as input, presumably due to turn-taking during social communication. See the supplementary material for further analysis.

\textbf{Result on Random Signals.}
As a baseline, we also perform an experiment by using signals from a random individual in the random sequences without any social link to our target individual, which shows about the chance level in Table~\ref{table:speaking_class_ablation}.

\begin{table}[t]
	\centering
	\footnotesize
	\begin{tabular}{c| c| c| c}
		\hline
		Input Signal & Self  & Other seller's & Random person's\\
		\hline
		Face+Body & 88.40\% & 78.42\% & 49.65\%\\       
		\hline
		Face & \underline {\textbf{88.93}}\% & \underline {\textbf{80.14}}\% & 49.64\%\\       
		\hline
		Body & 73.12\% & 70.48\% & 50.22\%\\       
		\hline \hline
		F+B, Masked Body & 82.48\% & 75.10 \% & -- \\        
		\hline
		F+B, Masked Face & 56.59\% & 64.27 \% & -- \\       
		\hline		
	\end{tabular}
	\caption{Speaking status classification accuracy using different social signal sources as input. The last two rows test performance of the "Face+Body" model after zeroing out parts of the input data without retraining.\label{table:speaking_class}\label{table:speaking_class_ablation}}
\end{table}
\begin{figure}[t]
	\centering       
	\includegraphics[ width=0.9\linewidth]{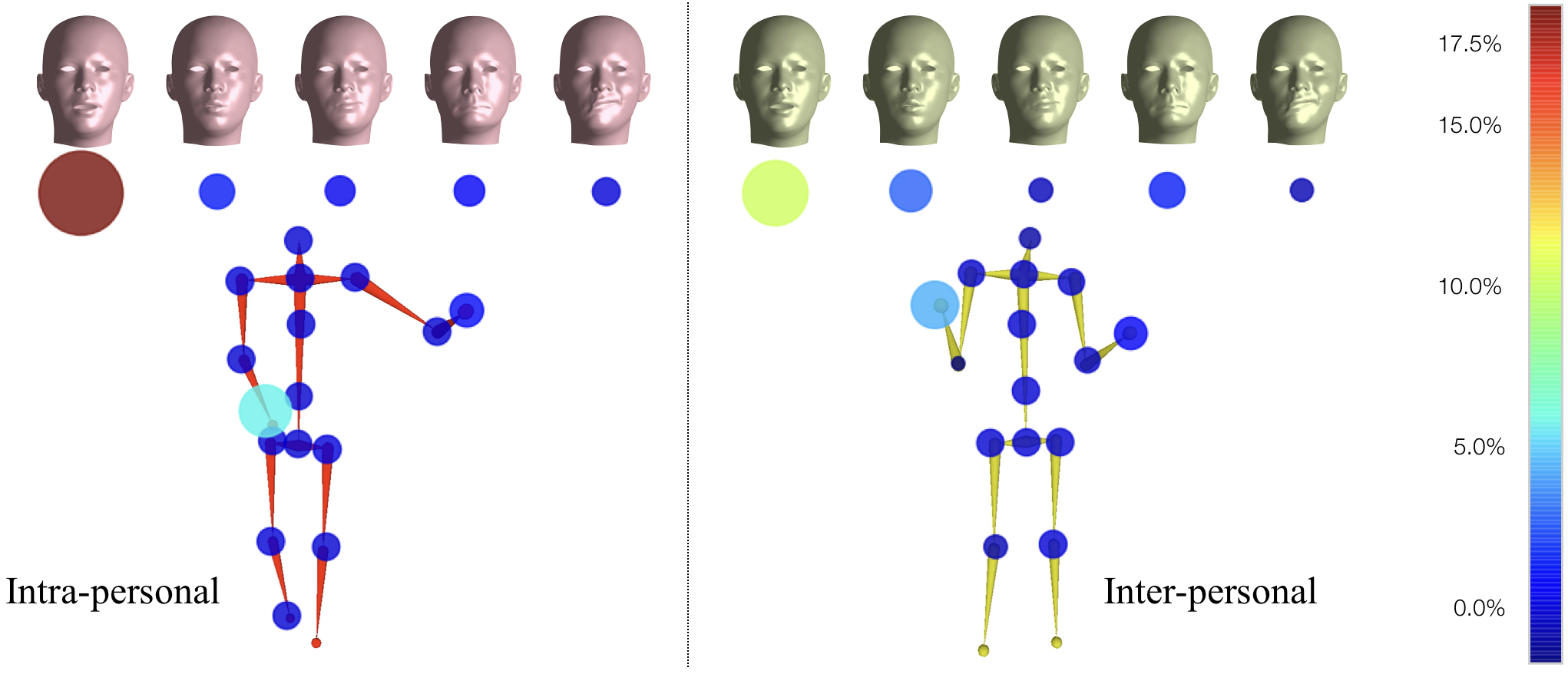}
	\caption{An ablation study by comparing the prediction performances after removing each channel of the social signal input from the trained networks of ``Face+Body''in Table~\ref{table:speaking_class}. The performance drops after removing each part, compared to the original performances, are shown by colors and circle sizes. The left figure is the result by using the target person's own signals, and the right figure is by using the other seller's signals. The colorbar on the right shows the frame drops in percentage from the original performances.} 
	\label{fig:result_speakPred_ablation}
\end{figure}


\textbf{An Ablation Study to Verify the Influence of Each Part.} We perform an ablation study by comparing the prediction performance after removing every single channel in the trained network. For this test, we use the network trained with ``Face+Body'' in Table~\ref{table:speaking_class}. We mask out a certain channel (e.g., a face motion component or a body joint part) in the input at test time and check the performance drop from the original output. The result is shown in Figure~\ref{fig:result_speakPred_ablation}, where the colors and the sizes of the circles represent the amount of performance decrease. This result shows that the first component of the face motion, which is corresponding to the mouth opening motion, has the strongest predictive power for speaking status. As another interesting result, the result shows that the right hand has stronger predictive power than the left hand.

\subsection{Social Formation Prediction}
We predict the position and orientation of the target person, the ``left seller", by using the signals of communication partners. In this test, we explore the prediction accuracy by considering combinations of difference sources: using body position, body orientation, and face orientations. Table~\ref{table:predForm_errors} shows the results. By using all signals, we obtain the best performance. Intuitively, we can imagine that the target person's location can be estimated by triangulating the face normal direction of the other two subjects, which presumably learned from our network. The prediction performance using only position cues shows the worst, but still a reasonable, performance among them. 


\begin{table}[t]
	\centering
	\footnotesize
	\begin{tabular}{c| c| c| c}
		\hline
		Types & Position (cm) & Body Ori. (\degree) & Face Ori. (\degree)\\
		\hline
		PosOnly & 29.83 (13.38) & 15.24 (7.23) & 19.02 (7.64) \\
		\hline
		Pos+face & 25.23 (9.74) & 13.20 (5.17) & 17.61 (6.89) \\
		\hline
		Pos+body & 26.57 (10.24) & 12.80 (4.37) & 17.51 (5.60) \\
		\hline
		Pose+face+body & \underline {\textbf{24.59}} (10.23) &  \underline {\textbf{12.33}} (3.71) &  \underline {\textbf{17.01}} (5.18) \\
		\hline
	\end{tabular}
	\caption{Social Formation Prediction Errors (Std.). Average position error between our estimation and ground-truth are reported in centimeters. The body and face orientation errors is in degree between the estimated facial/body normal direction and ground truth.\label{table:predForm_errors}}
\end{table}

\begin{table}[t]
	\centering
	\footnotesize
	
	\begin{tabular}{c| c| c}
		
		\hline
		Types & Avg. Joint Err. (cm) & Std.\\
		\hline
		$\mathcal{F}_{p\rightarrow B0 }$ (Eq.~\ref{eq:pred_traj2body}) & 8.31 & 2.26\\
		\hline
		$\mathcal{F}_{(B1,B2)\rightarrow B0}$ (Eq.~\ref{eq:pred_body2body}) & 8.72 & 2.00\\
		\hline
		Hybrid & 8.61  & 1.84\\
		\hline
		Average body (baseline) & \underline {\textbf{7.83}} & 2.33\\
		\hline
	\end{tabular}
	\caption{Social Body Gesture Prediction Errors (cm). Hybrid uses $\mathcal{F}_{p\rightarrow B0 }$ (Eq.~\ref{eq:pred_traj2body}) for lower body and $\mathcal{F}_{(B1,B2)\rightarrow B0}$ (Eq.~\ref{eq:pred_body2body}) for upper body prediction.\label{table:predBody_errors} }
\end{table}

\subsection{Body Gestures Prediction}
The quantitative evaluation of body gesture predictions is shown in Table~\ref{table:predBody_errors}. In the first method of Eq.~\ref{eq:pred_traj2body}, the body motion is directly regressed from the estimated social formation (location and orientation) of the target person, and the output shows reasonable leg motions following the trajectory movements, but has minimum upper body motion. As an alternative method, in the method of Eq.~\ref{eq:pred_body2body}, the output does not take into account the global formation information, but shows more dynamic and realistic body motions by responding to other subjects' motion. Finally, we combine both methods (labelled ``Hybrid''), by merging the formation and leg motion from the first method to the upper body motion from the second method, which generates qualitatively better results than others, satisfying most of the noticeable social rules in the scenes (distance, orientation, leg and root movement, and natural hand motions). However, the quantitative errors tend to be higher. Notably, the baseline method, always generating a fixed ``mean pose'' computed from the training set, shows the best performance. This is because the error metric computing the 3D errors from the ground-truth cannot fully evaluate how natural the motion appears. See the supplementary video.

\subsection{Discussion}
We present a data-driven social signal prediction framework, which allows us to investigate the dynamics and correlations among interpersonal social signals. We formalize the social signal prediction framework, and describe sub-tasks considering various channels of input and output social signals. To build the models, we collect the Haggling dataset from hundreds of participants, and demonstrate clear evidence that the social signals emerging in genuine interactions are predictive each other. The approach described in this paper is an important direction to endow machines with nonverbal communication ability. There are still several unexplored issues, including how to better evaluate more natural behaviors, modeling both verbal and nonverbal signals together, and modeling more diverse social interactions than triadic scenarios.
\vspace{2mm}

\noindent \textbf{Acknowledgements.} We thank Hyun Soo Park, Luona Yang, and Donielle Goldinger for their help and discussions in designing and performing the data capture.

{\small
\bibliographystyle{ieee}
\bibliography{thesis}
}

\clearpage

\appendix

{\Large \textbf{Appendix.}}

\section{The Haggling Game Protocol}
To evoke natural interactions, we involved participants in a social game named the \emph{Haggling} game. We invent this game to simulate a haggling situation among two sellers and a buyer. The triadic interaction is chosen to include interesting social behaviors such as turn taking and attention changes, which are missing in previous dyadic interaction datasets~\cite{rehg2013decoding}. During the game, two sellers are promoting their own comparable products for selling, and a buyer makes a decision about which product he/she buys between the two. The game lasts for a minute, and the seller who has sold his/her product is awarded $\$5$. To maximize the influence of each seller's behavior on the buyer's decision-making, the items assigned to sellers are similar products with slightly different properties. Example items are shown in Table~\ref{table:haggling_items}. 

\begin{figure}
	\centering
	\includegraphics[trim=0 0 0 0,clip,width=\linewidth]{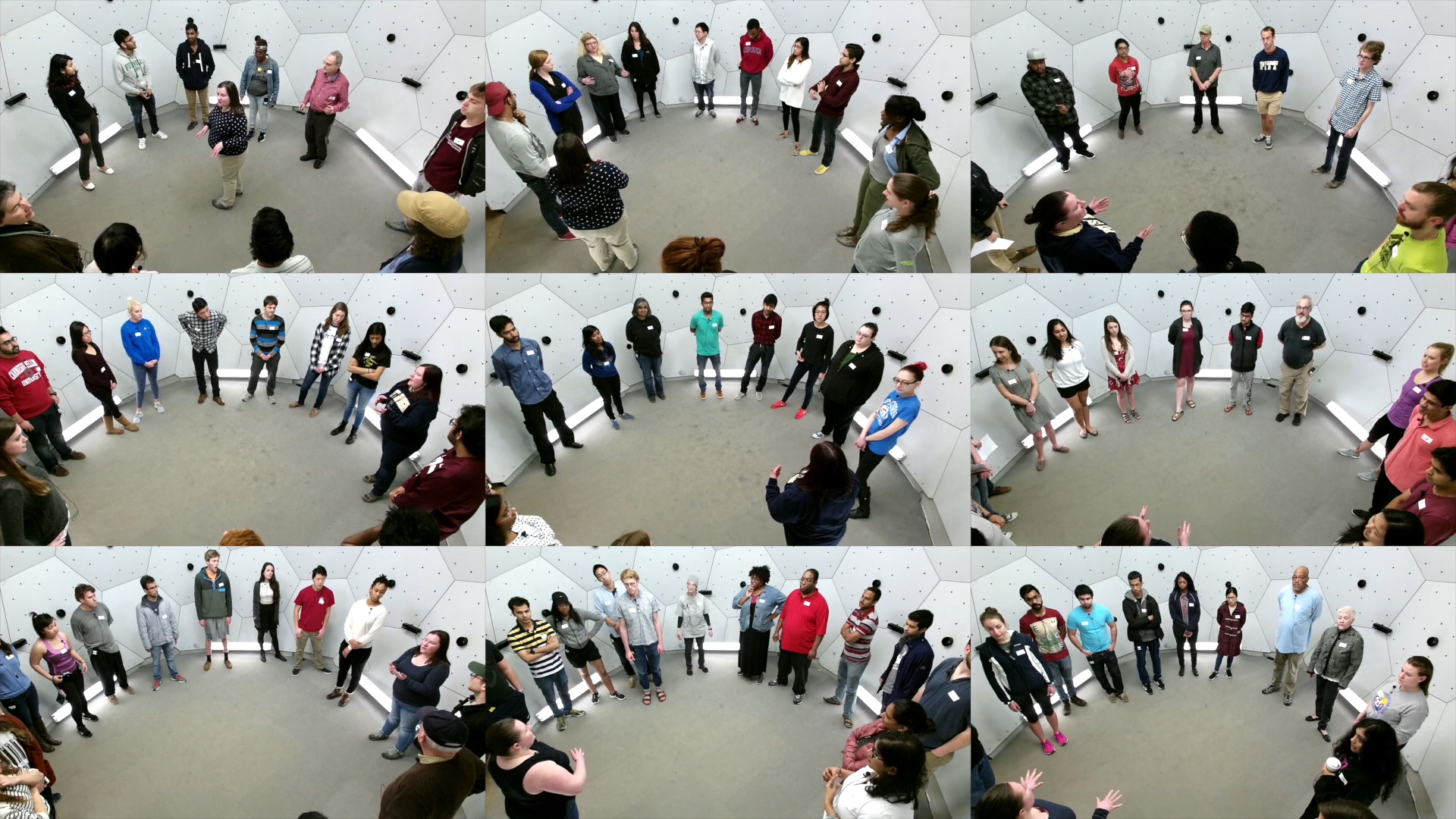}
	\caption{Before starting the social game capture, participants are instructed the game rules and also spent time to be accustomed to the Panoptic Studio environment, as shown in these photos. We follow a common and strict protocol during all captures to avoid any potential bias.} 
	\label{fig:haggling_intro}
\end{figure}

\begin{figure*}
	\centering
	\includegraphics[width=\textwidth]{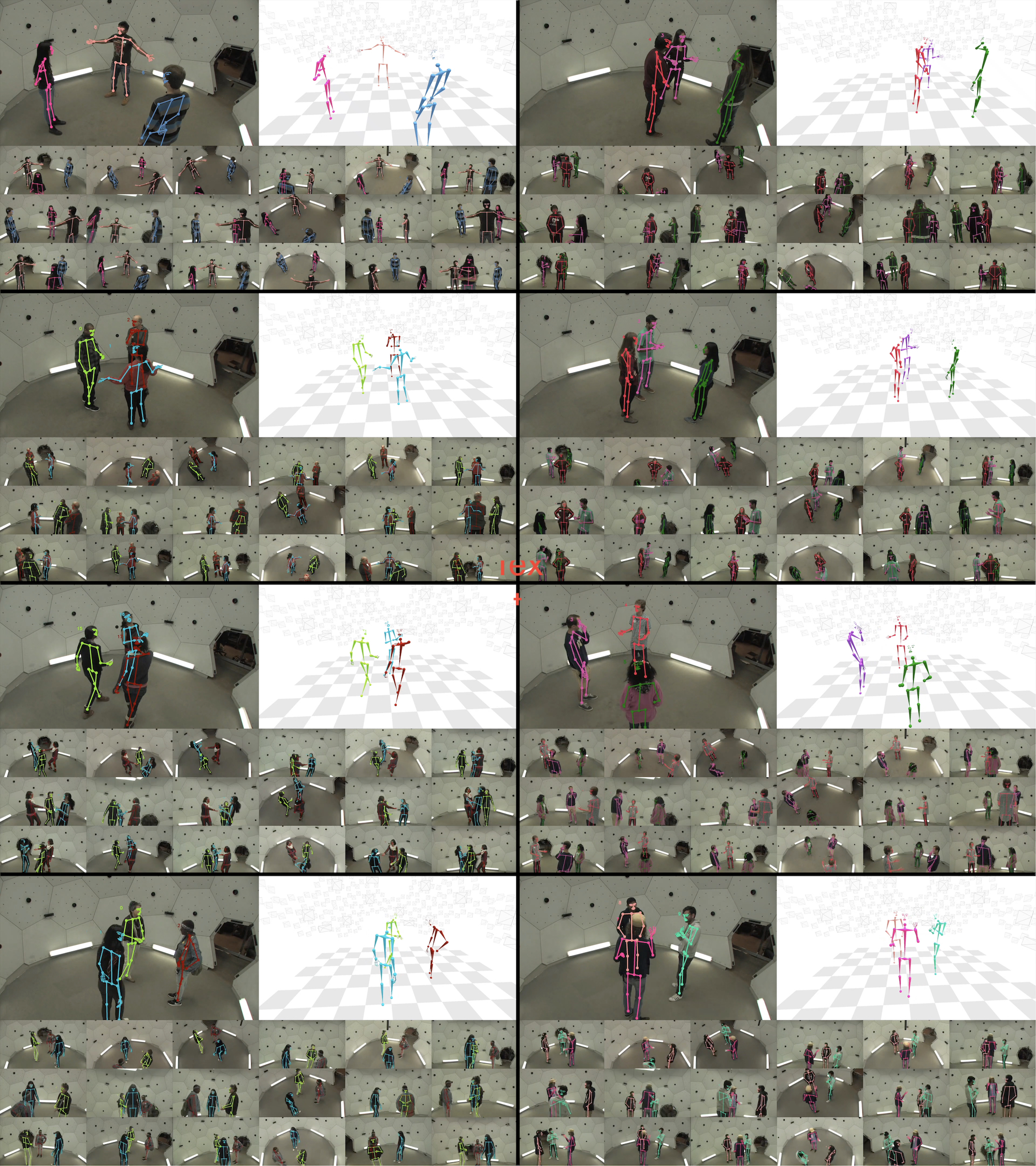}
	\caption{Example scenes of haggling sequences with social signal measurements. For each example, HD images overlaid by the projections of 3D anatomical keypoints (from bodies, faces, and hands) are shown, along with a 3D view of the social signal measurements (top right).} 
	\label{fig:haggling_db}
\end{figure*}

For every capture, we follow the protocol described below. We randomly recruited participants using the CMU Participant Pool\footnote{\url{https://cbdr.cmu.edu/}}. Over the 8 days of captures, 122 subjects participated and 180 haggling sequences were captured (about 3 hours of data). The participants arrive at the lab for the capture and first sign the IRB consent form with an agreement to publicly release the data for research purposes only. A unique identification number is assigned to each participant, and participants are also equipped with a wireless microphone. Then, all subjects are informed of the rules of the Haggling game by watching a pre-recorded presentation video together. Notably, they are not instructed about how to behave during the game, nor is their clothing or appearance controlled. All motions in the sequences are spontaneous social behaviors based on the informed game rules. After introducing the game rules, participants are asked to spend time inside the studio (as shown in Figure~\ref{fig:haggling_intro}) so that they can be accustomed to the interior view of the Panoptic Studio~\cite{joo2015panoptic, joo2017panoptic}. Before starting the capture, groups and roles are randomly assigned, and participants line up based on their numerical orders. We provide written descriptions to sellers about the items they will be selling in small cards 1 minute before the game, and the sellers return the card before entering the studio. With a starting signal, participants in a group enter the studio and start the haggling game immediately. The positions and orientations of the groups inside the system are also spontaneously decided (no instructions are given). During the capture, their social signals including voice, positions, orientations, and body motions are recorded.  We send a signal by ringing a bell 10 seconds before the end of the game, and send the same alarm at the end of the game. After the capture, the buyer annotates the decision between the two items in the prepared result sheet. The captured sequences contain many voluntary social behaviors of diverse people in a common social context. Example scenes are shown in Figure~\ref{fig:haggling_db} and our supplementary video.

\begin{table}[h]
	\centering
	\begin{tabular}{ l| l | l }
		\hline	
		Items & Seller 1  &  Seller 2\\
		\hline	
		\hline	
		Phone &  \makecell{Light weight \\ Medium storage}    & \makecell{Medium weight\\ Large storage}\\
		\hline			
		Laptop &  \makecell{Light weight \\ Medium speed}    & \makecell{Medium weight\\ Fast speed}\\
		\hline			
		Tablet PC &  \makecell{Large storage\\ Medium speed}    & \makecell{Medium storage\\ Fast speed}\\
		\hline	
		Speaker &  \makecell{High quality audio\\ Wired}    & \makecell{Medium quality audio\\ Wireless}\\
		\hline	
	\end{tabular}
	\caption{Examples of items assigned to sellers in our Haggling games.}
	\label{table:haggling_items}
\end{table}

\section{Face Motion Parameters}
\begin{figure}[t]
	\includegraphics[trim=10 10 10 10,clip,width=\columnwidth]{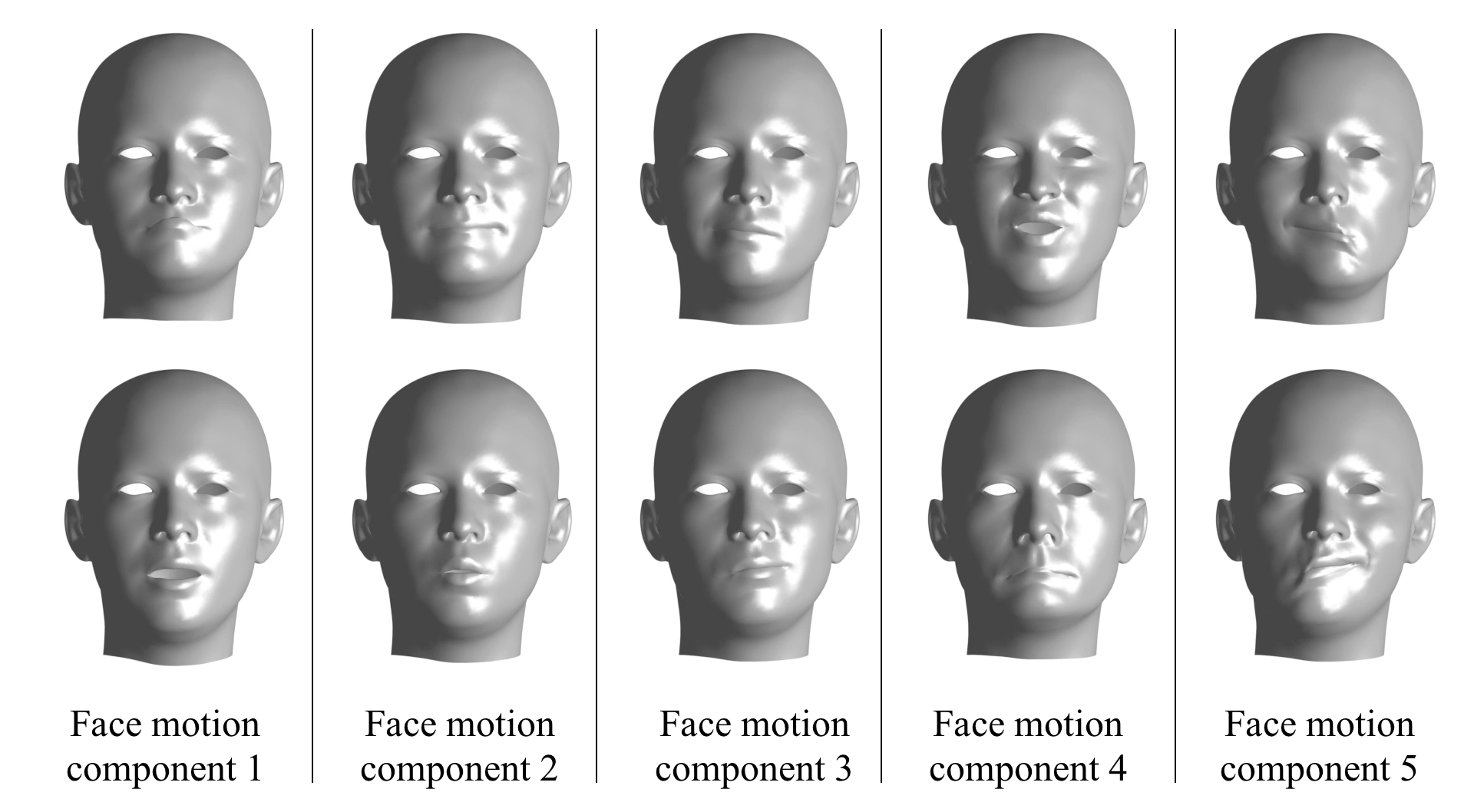}
	\caption{The five face motion components (showing parameter weights with -0.3 on the top and 0.3 on the bottom) used in our social signal modeling.}
	\label{fig:notation_input}
\end{figure}

For the face motion signal, we first fit the deformable face model of \cite{cao2014facewarehouse} and use the initial 5 dimensions of the facial expression parameters, because we found the remaining dimensions have an almost negligible impact on our reconstruction quality. Note that the face expression parameters in \cite{cao2014facewarehouse} are sorted by their influence by construction and the initial components have more impact in expressing facial motion. To this end, face motion at a time instance is represented by a 5-dimensional vector, $\mathbf{F}(t) \in \mathbb{R}^{5}$, and example facial expressions expressed by each component are shown in Figure~\ref{fig:notation_input}.

\section{Implementation Details of Social Signal Predictions}
In this section, we discuss the details of our neural network architectures to implement the social signal prediction models for each sub-task.

\begin{figure}[t]
	\includegraphics[width=\columnwidth]{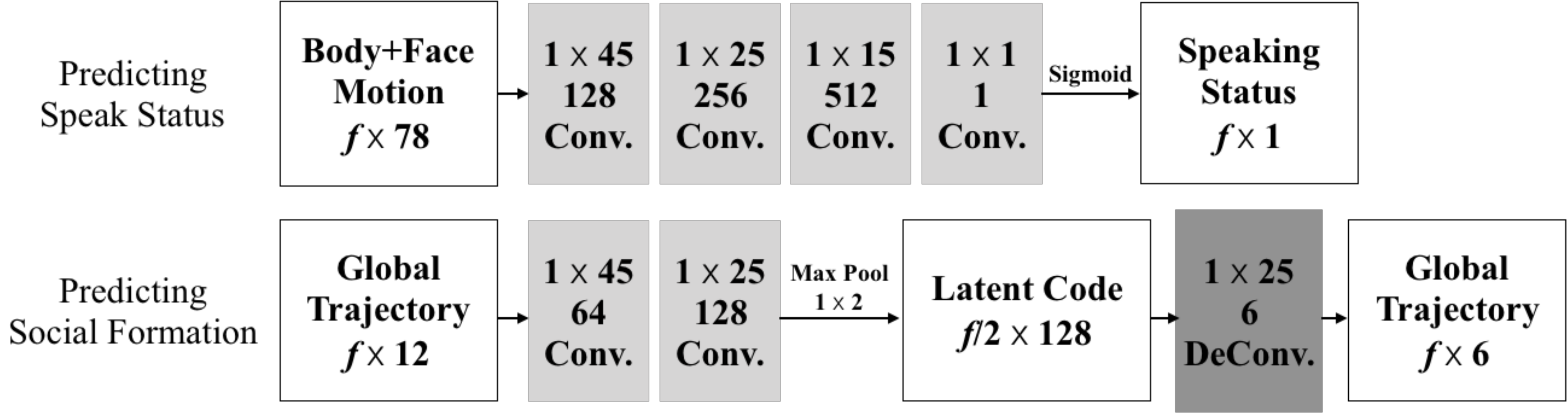}
	\caption{Network architectures for predicting speaking status and predicting social formation. We use fully convolutional networks for both sub-tasks.}
	\label{fig:ssp_imp_detail}
\end{figure}

\subsection{Predicting Speaking}
Our neural network is composed of four 1-D convolutional layers (see Figure~\ref{fig:ssp_imp_detail}). The first three layers output 128, 256, and 512 dimensional features respectively with ReLU activation functions, and the last layer has $1\times1$ convolutions with a sigmoid activation layer. Dropout~\cite{srivastava2014dropout} is also applied for the second and third layers with a probability of 0.25. Our model does not require a fixed window size for the input (since it is fully convolutional), but we separate input data into small clips with a fixed size (denoted by $f$) for efficiency during training. During testing time, our models can be applied to the input of arbitrary length. We use $f=120$ (4 seconds) for the input window size for training, and we use an arbitrary length of input data for testing. The feature dimension of the input to our network is the concatenation of face motion and body motion (78 dimensions). If fewer cues are used (e.g., face only or body only), we mask out the unused channels as their average values computed in the training set and keep the same network structure. We use an adaptive gradient descent algorithm, AmsGrad algorithm~\cite{reddi2018convergence}, implemented in PyTorch~\cite{paszke2017automatic}, along with $l^1$ regularization loss with 0.001 as the regularization strength. 

\subsection{Predicting Social Formations}
Our neural network has an autoencoder structure, where the encoder is composed of two 1-D convolutional layers followed by a max pooling layer with stride 2, and the decoder is composed of a single 1-D transposed convolution layer (see Figure~\ref{fig:ssp_imp_detail}). The output feature dimensions are 64, 128, and 6 respectively. Dropout~\cite{srivastava2014dropout} is also applied in front of all layers with a probability of 0.25. Similar to the speaking status prediction, our model does not require a fixed window size for the input, but we separate input data into small clips with a fixed size ($f=120$, or 4 seconds) for the efficiency in training. The input is the concatenation of the cues of the other two communication partners (12 dimensions) with a fixed order (buyer and then the right seller), and the output of our network is the position and orientations of the target individual, the left seller (6 dimensions). Similar to the previous prediction task, if fewer cues are used (e.g., position only), we mask out the unused channels as their average values computed in the training set and keep the same network structure. We use an adaptive gradient descent algorithm, AmsGrad algorithm~\cite{reddi2018convergence}, implemented in PyTorch~\cite{paszke2017automatic}, along with $l^1$ regularization loss with 0.1 as the regularization strength.

\subsection{Predicting Body Gestures (Kinesic Signals)}
\begin{figure}[t]	
	\includegraphics[width=\columnwidth]{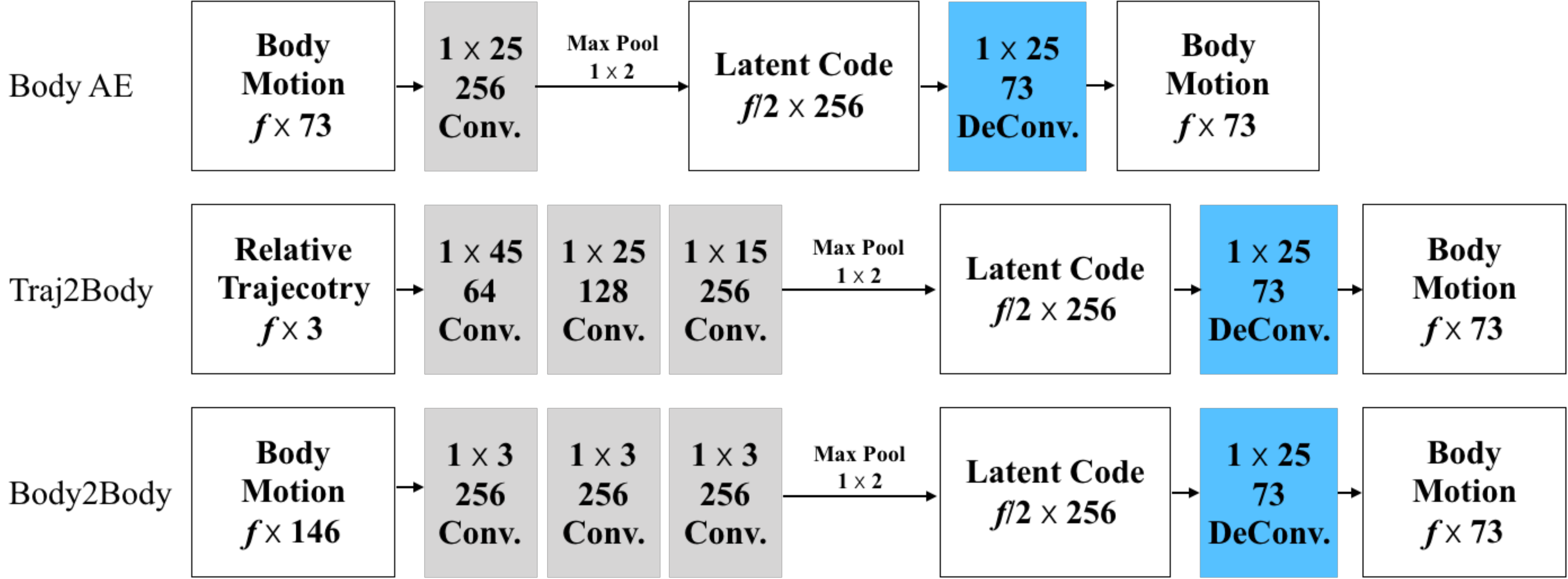}
	\caption{Network architectures for body gesture prediction. We adopt the autoencoder architecture of Holden et al.~\cite{holden2016deep} to learn the latent space for by motion shown in the first row. We consider two different approaches to generate the latent code, from the predicted social formation of the target individual or from the communication partners' body motion. Note that the decoder part shown as the blue boxes are frozen after training the body motion autoencoder.}
	\label{fig:body_network}
\end{figure}

As in the work of Holden et al.~\cite{holden2016deep}, we first train an body motion autoencoder (as shown in the first row of Figure~\ref{fig:body_network}) to find the motion manifold space, so that the decoded output from the latent space can express a reasonable human body motion. Then, we keep the decoder part of this network (shown as the blue boxes in Figure~\ref{fig:body_network}) for the gesture prediction, which uses the latent codes generated by the following two different approaches as input. 

\paragraph{From the Body Trajectory:}
We regress the latent code for the gesture prediction from the estimated trajectory information of the target person (position and body orientation). The network architecture is shown in the second row of Figure~\ref{fig:body_network}. Note that we freeze and do not train the decoder part (the blue box) which is taken from the body motion autoencoder. As input, the model uses the velocities of position and body orientation (relative root movements with respect to the previous frame), which is a subpart of our body motion representation (the first 3-dimensions out of the 73-dimensional vector). For training, we use ground truth body motion data, by using the subpart representing relative root movements as input, and all dimensions for body motion as output. During testing, we convert the social formation prediction output (global position and orientation) into this velocity representation (relative position and orientation), and use it as the input for this network. 

\paragraph{From Other Body Gestures:}
In this case, we use the other two partners' body motions as input to generate the latent code, and decode it to predict the target individual's body gesture, similar to the previous approach. The network architecture is shown in the third row of Figure~\ref{fig:body_network}.

\section{Revisiting Proxemics}
\begin{figure}
	\centering       
	\includegraphics[ width=\linewidth]{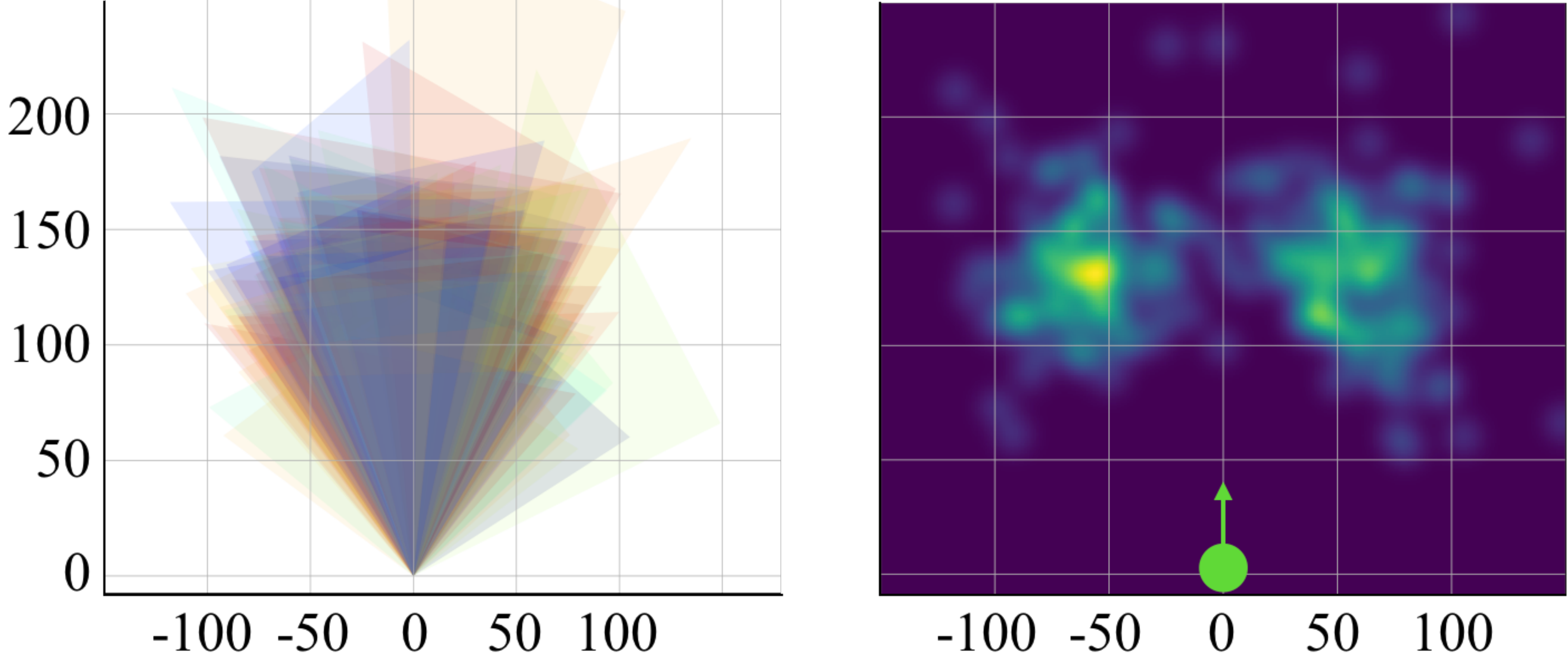}
	\caption{Visualizing social formations in the haggling sequences as triangles (left) and a heat map (right). The formation is normalized w.r.t the buyer's location, and the green circle on the right shows the buyer location (origin) and orientation ($z$-axis).} 
	\label{fig:socialgeo_distribution}
\end{figure}

Our dataset has the measurement of fully spontaneous motions (including the position and orientation of groups) of interacting people, and enables us to revisit the well-known proxemics theory~\cite{Hall66}. We first compute the average, minimum, and maximum distances between a pair of subjects: (1) buyer and right sellers (B-RS), (2) buyer and left seller (B-LS), and (3) left seller and right seller (LS-RS). The results are shown in Table~\ref{table:proxemics_comp}. We found that the result approximately follows the social distance categories defined in the Hall's categorization~\cite{Hall66}. The average distances among subjects are within the close phase of social distance ranges (from 120 $cm$ to 210 $cm$) and the max distances are within the far phase of social distance (from 210 $cm$ to 370 $cm$) in \cite{Hall66}. To analyze the shape of the social formation, we plot the average formation of games in a person-centric coordinate by a buyer. The results are shown in the Figure~\ref{fig:socialgeo_distribution}, showing that the formation is often similar to isosceles triangles with relatively far distances between a buyer and two sellers than the distance between sellers. 

\begin{table}[t]
	\centering
	\caption{Average distances (cm) between subjects. B, RS, and LS denote buyer, right seller, and left seller respectively.}
	\label{table:proxemics_comp}
	\begin{tabular}{l| l| l| l| l}
		\hline
		& Avg. dist. & Std. & Min & Max \\
		\hline
		B-RS & 148.11 & 27.26 & 99.03 & 265.52 \\
		\hline
		B-LS & 151.45 & 29.62 & 104.24  & 284.85 \\
		\hline
		LS-RS & 124.13 & 24.05  & 77.70  & 206.26 \\
		\hline
	\end{tabular}
\end{table}

\begin{figure}[t]
	\centering       
	\includegraphics[ width=0.9\linewidth]{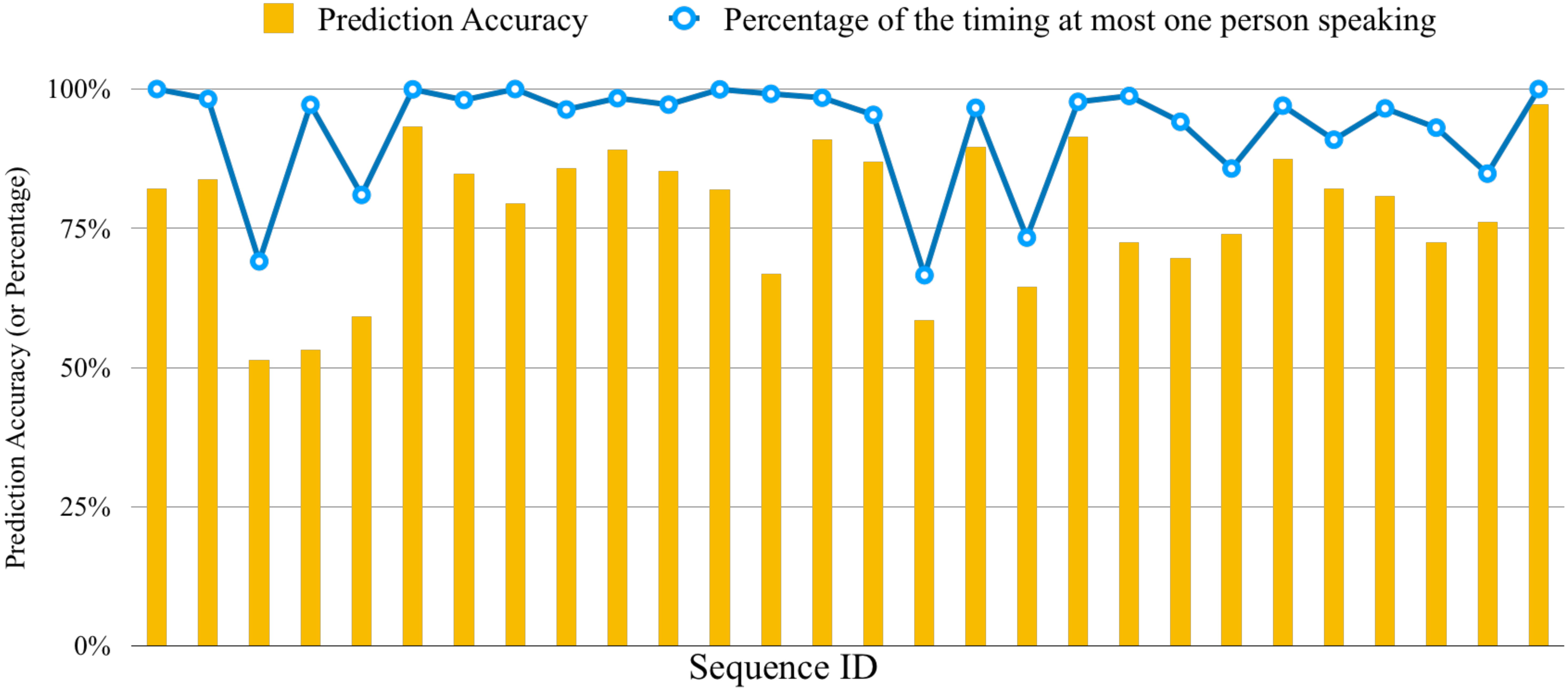}
	\caption{Comparison between the performance of speaking status prediction and turn-taking status for each sequence. Each column shows the prediction performance (yellow bar) where the other seller's face and body signals are used as input. The blue curve represents how well the turn-taking rule is satisfied, which is defined by counting the percentage of the timing where at most one person is speaking. } 
	\label{fig:turn-taking}
\end{figure}

\begin{figure*}[t]
	\centering       
	\includegraphics[trim=20 10 2 10,clip, width=\linewidth]{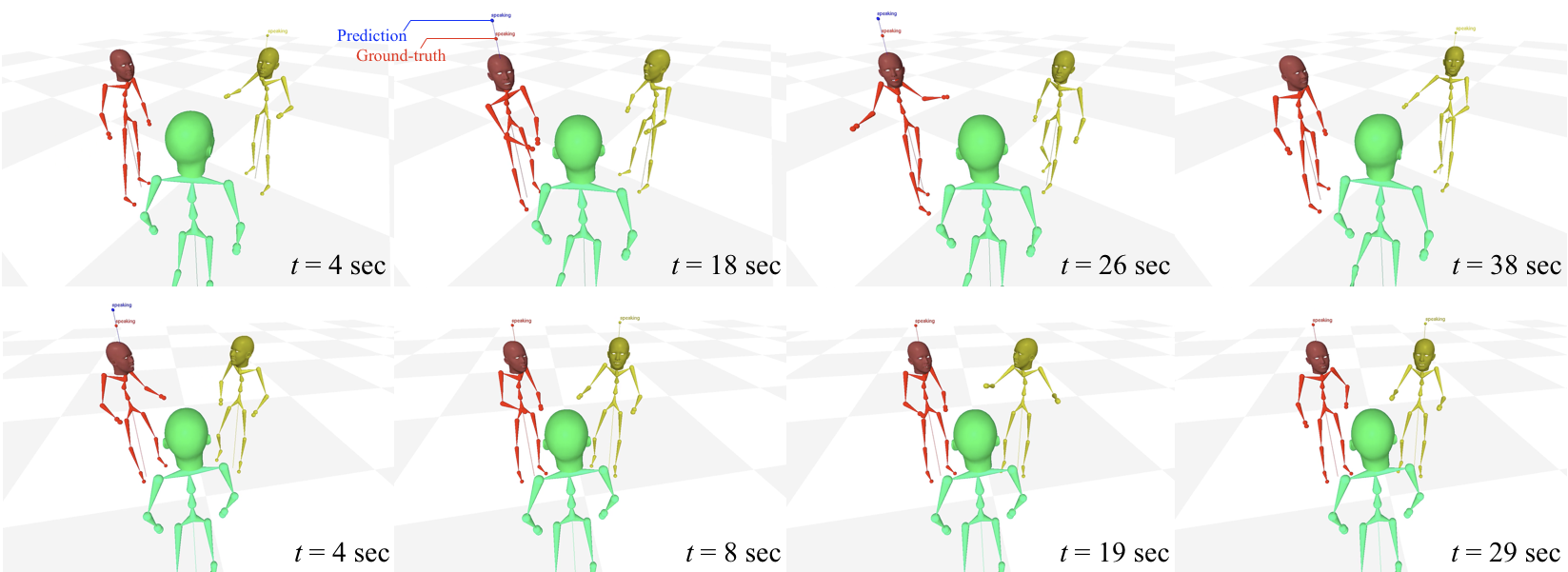}
	\caption{Qualitative results of the speaking prediction of the target person (red) by using the other seller's (yellow) face and body motions as input. The speaking prediction output is shown as the blue ``speaking" label above the target person's head, while the ground truth speaking status is shown as the red label. The prediction is accurate, if both blue and red labels are shown or not shown. Examples scenes of two haggling games (top and bottom) are shown, where the sequence on the top has high accuracy (89\%) and the sequence on the bottom has low accuracy (58\%). In the haggling game shown on the top, both sellers follow turn taking almost always, while both sellers frequently speak at the same time in the sequence shown on the bottom.} 
	\label{fig:qual_spek_pred}
\end{figure*}

\begin{figure*}[t]
	\centering       
	\includegraphics[trim=20 2 2 10,clip, width=\linewidth]{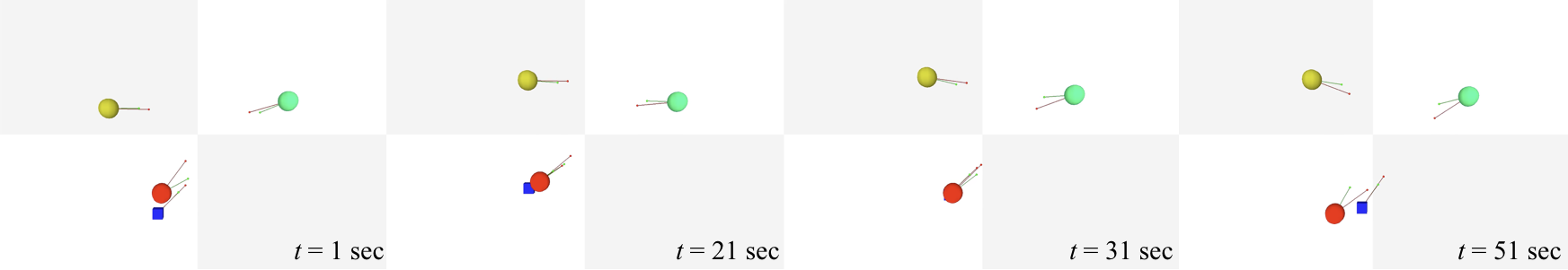}
	\caption{Qualitative results of the social formation prediction of a haggling game, visualized from the view at the top. The target person is shown as red spheres. The cues from other people (yellow and cyan spheres) are used as input for the prediction, and the prediction output is shown as the blue cube. The red lines represent body orientations, and the green lines represent face orientations.} 
	\label{fig:qual_form_pred}
\end{figure*}

\section{Further Analysis on Speaking Status Prediction}

\paragraph{Result on Inter-personal Signals.}
As shown in the second column (other seller's input) of Table 1 of the paper, the result clearly shows that there exists a strong link between interpersonal social signals. The other seller's facial motion shows a strong predictive power for the target person's speaking status, where the accuracy is higher than the case of using the target person's own body signals as input, presumably due to the turn-taking property in social communication. For example, we can assume that the target person is not speaking, when the other seller is speaking. We can further investigate this by checking how well the turn-taking rule is satisfied during each social game scene, along with its predicting performance. As a way to measure the turn-taking status, we consider the percentage of the timing at which at most one person speaks, which defined by:
\begin{gather}	
\frac{\sum \limits_{t} \delta \left(\mathbf{S}^0(t) + \mathbf{S}^1(t) <2 \right)}{T},
\end{gather}
where $T$ is the total time of a Haggling game, $\mathbf{S}$ is the speaking status for sellers, and $\delta$ is a function that returns 1 if the condition satisfies and returns 0 otherwise. In this measurement, 100\% means that there is no time that both sellers are speaking at the same time, where the turn-taking rules are perfectly satisfied. We compute this measurement to check the turn-taking status for each testing sequence as the blue curve in Figure~\ref{fig:turn-taking}. In this figure, we also plot the speaking prediction accuracy for each testing sequence by using the other seller's both face and body signals as input, which is shown as yellow bars. As shown in the figure, the prediction performance shows a very similar pattern to this turn-taking status, and this means that this implicit social ``rule'' is a source of linking the social signals across individuals. Example qualitative results are shown in the Figure~\ref{fig:qual_spek_pred}.

\section{Qualitative Results}
Example results of speaking classification and social formation prediction are shown in Figure~\ref{fig:qual_spek_pred} and Figure~\ref{fig:qual_form_pred}. Results are best seen in the supplementary videos.

\end{document}